\documentclass{article} 
\usepackage{iclr2027_conference,times}

\usepackage{amsmath,amsfonts,bm}

\def\eqref#1{equation~\ref{#1}}

\def\1{\bm{1}}

\DeclareMathAlphabet{\mathsfit}{\encodingdefault}{\sfdefault}{m}{sl}
\SetMathAlphabet{\mathsfit}{bold}{\encodingdefault}{\sfdefault}{bx}{n}

\usepackage{graphicx}
\usepackage{booktabs}
\usepackage{capt-of}
\usepackage{algorithm}
\usepackage{algpseudocode}
\usepackage[most]{tcolorbox}
\usepackage{listings}
\usepackage{xcolor}
\usepackage{colortbl}
\usepackage{hyperref}
\usepackage{url}

\title{CoEvolve: Counterfactual-Guided\\ Co-Evolution of Memory and Collaboration\\ in Multi-Agent Systems}

\author{
\textbf{
Xufei Shi$^{1}$,
Derong Xu$^{1}$,
Wei Chen$^{1}$,
Yanyu Yao$^{1}$,
Zhi Zheng$^{1}$,
Tong Xu$^{1}$
}
\\
$^{1}$State Key Laboratory of Cognitive Intelligence,\\
University of Science and Technology of China, Hefei, China\\
{\footnotesize
\texttt{\{sxf2022, derongxu, chenweicw, yyyao\}@mail.ustc.edu.cn}\\
\texttt{\{zhengzhi97, tongxu\}@ustc.edu.cn}
}
}

\iclrfinalcopy

\begin{document}

\maketitle

\begin{abstract}
Large language model (LLM)-based multi-agent systems have demonstrated remarkable capabilities across diverse tasks. Yet their memory and collaboration architectures often remain static, constraining continued adaptation and improvement over time. Recent work addresses this limitation by evolving memory or collaboration based on interaction experience. However, updates to these interdependent components are not necessarily aligned: an evolved workflow may require experience that memory does not provide, while evolved memory may encode guidance that the workflow cannot use. To address this problem, we propose \mbox{CoEvolve}, a skill-centered framework for aligned memory--collaboration evolution. \mbox{CoEvolve} represents reusable local collaboration patterns as skills that link evolving collaboration decisions to their supporting memories. It further compares original and counterfactual trajectories to identify local evolution signals, which coordinate updates to the skill library, skill memory, and executor memories across rounds. Experiments on the ALFWorld and WebShop benchmarks confirm that CoEvolve achieves the best overall performance among the evaluated baselines, while ablation studies further validate the effectiveness of aligned co-evolution and counterfactual-guided evolution. Our code is available at \url{https://github.com/MINE-USTC/CoEvolve}.
\end{abstract}

\section{Introduction}
\label{sec:introduction}

Large language model (LLM)-based agent systems have demonstrated strong capabilities on complex reasoning tasks and are increasingly deployed in open-ended settings that require long-horizon planning, collaboration, and sustained interaction with the environment~\citep{hao2026evolveteam,zhang2026evochamber}. Yet the underlying models are typically fixed after training, while the memory and collaboration architectures of many multi-agent systems (MAS) remain unchanged after deployment~\citep{gao2026survey}. Consequently, these systems have limited ability to incorporate interaction experience and adapt to new tasks, evolving knowledge, or changing environments.

To move beyond static designs, recent self-evolving approaches use task outcomes or execution traces to revise persistent MAS components across rounds. One line of work evolves how memory represents, organizes, and retrieves experience. For example, MemEvolve~\citep{zhang2025memevolve} jointly evolves experiential knowledge and its modular memory architecture, while MemSkill~\citep{zhang2026memskill} evolves reusable memory operations and their selection. Another line adapts how agents communicate and organize their workflows. GPTSwarm~\citep{zhuge2024gptswarm}, for instance, optimizes communication graphs, whereas HFlow~\citep{liu2026hflow} revises workflow structures using execution experience and human--agent feedback. EvoChamber~\citep{zhang2026evochamber} further adapts both individual agents and team organization at test time. Together, these studies show that persistent components can improve from experience. However, memory and collaboration should not evolve independently because they are mutually dependent. An update to one component can change what the other needs to provide or consume.

\begin{figure}[t]
    \centering
    \includegraphics[width=0.96\linewidth]{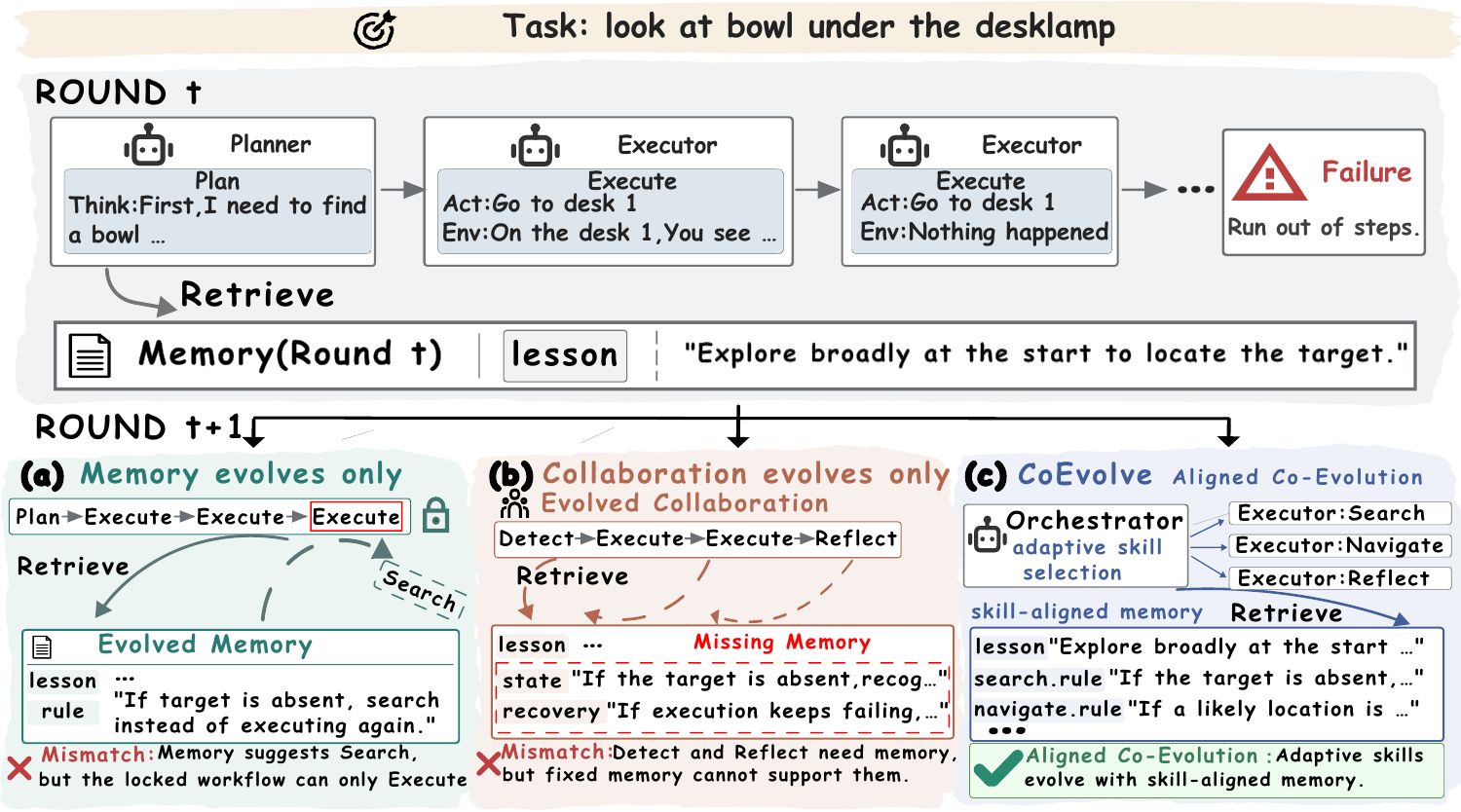}
    \caption{Cross-component mismatch under independent evolution: (a) memory-only and (b) collaboration-only updates leave workflow requirements and memory support misaligned, whereas (c) \mbox{CoEvolve} aligns both via shared skills.}
    \label{fig:cross_component_mismatch}
\end{figure}

When this dependency is not maintained, a locally useful update may fail to improve overall system behavior. We call this incompatibility \emph{cross-component mismatch}. As shown in Figure~\ref{fig:cross_component_mismatch}(a), memory evolution may produce guidance that a fixed workflow cannot invoke. Conversely, Figure~\ref{fig:cross_component_mismatch}(b) shows that collaboration evolution may introduce decisions without sufficient experience in fixed memory. In both cases, collaboration requirements and memory support evolve out of sync. Figure~\ref{fig:cross_component_mismatch}(c) illustrates the desired case, where each collaboration capability remains aligned with the memory support required for its decisions. This motivates a shared interface that jointly coordinates the evolution of collaboration and memory.

Effective coordination also depends on identifying where updates should occur. Existing evolution signals provide two main forms of evidence for this decision. Performance-driven methods use task-level scores, such as pass rates or final rewards, to search for or select candidate agent systems~\citep{hu2024adas,zhuge2024gptswarm,zhang2024aflow}. While these scores enable comparisons between system designs, they provide limited guidance on which local decisions should be improved. In contrast, process-driven methods inspect execution experience to diagnose potential improvements. ReCreate~\citep{hao2026recreate} reasons over interaction histories before editing agent scaffolds; Meta-Team~\citep{hao2026evolveteam} organizes distributed evidence through post-task communication; and RetroAgent~\citep{zhang2026retroagent} uses retrospective feedback to refine agent behavior. Although more informative, these signals remain grounded in the observed trajectory and do not directly test whether replacing a local segment would change the outcome. Decision-level counterfactual credit assignment provides a more direct test by evaluating local alternatives under matched context~\citep{chen2026exact}. However, identifying an outcome-sensitive segment alone does not indicate how the corresponding collaboration pattern and memory support should evolve together.

Building on these analyses, we propose \mbox{CoEvolve}, a counterfactual-guided framework for the aligned co-evolution of memory and collaboration. Keeping the two evolution processes aligned requires a shared unit for local collaboration capabilities and their supporting experience. To this end, \mbox{CoEvolve} represents reusable local collaboration patterns as skills and organizes three evolvable components around them: a skill library stores executable collaboration patterns, skill memory guides stage allocation, and executor memories support execution. This design supplies new collaboration capabilities with relevant memory while giving evolved memory concrete decision interfaces. To obtain evolution signals that indicate where updates should occur, \mbox{CoEvolve} constructs counterfactual trajectories from matched branch points after each round and compares them with the original trajectories. The resulting differences are further localized to identify outcome-sensitive pivot segments. These pivot-level signals jointly update the skill library, skill memory, and executor memories, keeping collaboration capabilities aligned with their memory support across rounds.

Our main contributions are summarized as follows:
\begin{itemize}
    \item We identify cross-component mismatch as an alignment challenge in self-evolving MAS and propose \mbox{CoEvolve}, which uses skills as shared interfaces to coordinate the evolution of collaboration patterns and their supporting memories.
    \item We introduce a counterfactual signal-acquisition method that identifies outcome-sensitive pivot segments and converts them into coordinated updates to the skill library, skill memory, and executor memories.
    \item We evaluate \mbox{CoEvolve} on ALFWorld and WebShop, where it achieves the highest overall scores among the evaluated methods, and ablation studies validate the effectiveness of aligned co-evolution and counterfactual-guided evolution.
\end{itemize}

\section{Related Work}
\label{sec:related_work}

\subsection{Self-Evolving Agents}
\label{sec:rw_selfevolving}

Self-evolving agents adapt based on feedback from their own interactions and execution, rather than relying solely on knowledge fixed at training time. This adaptation is carried by persistent components that survive across tasks. Early work evolves an artifact as a whole through evolutionary search, without separating its parts: EvoPrompt~\citep{guo2023evoprompt} evolves prompts with evolutionary operators; FunSearch~\citep{romera2024funsearch} discovers new programs by pairing a language model with an evolutionary loop; and AlphaEvolve~\citep{novikov2025alphaevolve} iteratively improves algorithms through direct code changes. Later work modularizes evolution by targeting a specific component: on the memory side, MemSkill, MemEvolve, and EvolveMem respectively evolve memory operations, modular memory architectures, and retrieval configurations~\citep{zhang2026memskill,zhang2025memevolve,liu2026evolvemem}; on the collaboration side, HFlow and Meta-Team revise workflow structures and team organization from execution evidence~\citep{liu2026hflow,hao2026evolveteam}. Recent studies also coordinate evolution across multiple components~\citep{zhang2026evochamber,cheng2026mem2evolve,wu2026cosplay}, but do not explicitly align evolving collaboration decisions with the memories that support them. \mbox{CoEvolve} targets this cross-component mismatch by treating skills as a shared interface and updating collaboration patterns together with their supporting memories.

\subsection{Agent Self-Evolution Methods}
\label{sec:rw_counterfactual}

Evolution signals have shifted from task-level outcomes to per-step evidence. Early methods treat the outcome of the entire system as the signal: ADAS~\citep{hu2024adas} retains the agentic systems proposed by its meta agent that score higher on the task; AFlow~\citep{zhang2024aflow} searches over workflow candidates with Monte Carlo tree search and uses execution results to guide the search; and GPTSwarm~\citep{zhuge2024gptswarm} optimizes node prompts and graph edges against downstream task performance. A second group derives per-step signals from execution experience: ReCreate~\citep{hao2026recreate} reasons over interaction histories before editing agent scaffolds; Meta-Team~\citep{hao2026evolveteam} organizes evidence through post-task communication; RetroAgent~\citep{zhang2026retroagent} uses retrospective numerical and linguistic feedback; and tool-interactive critique~\citep{gou2023critic} revises behavior based on verification feedback. Counterfactual methods compare local alternatives under matched context~\citep{chen2026exact} or verify critical steps by testing whether an alternative changes the outcome~\citep{li2026critical}. \mbox{CoEvolve} uses outcome-sensitive pivots to update a collaboration pattern together with the memory that supports it.

\section{Method}
\label{sec:method}

\begin{figure}[t]
    \centering
    \includegraphics[width=0.96\linewidth]{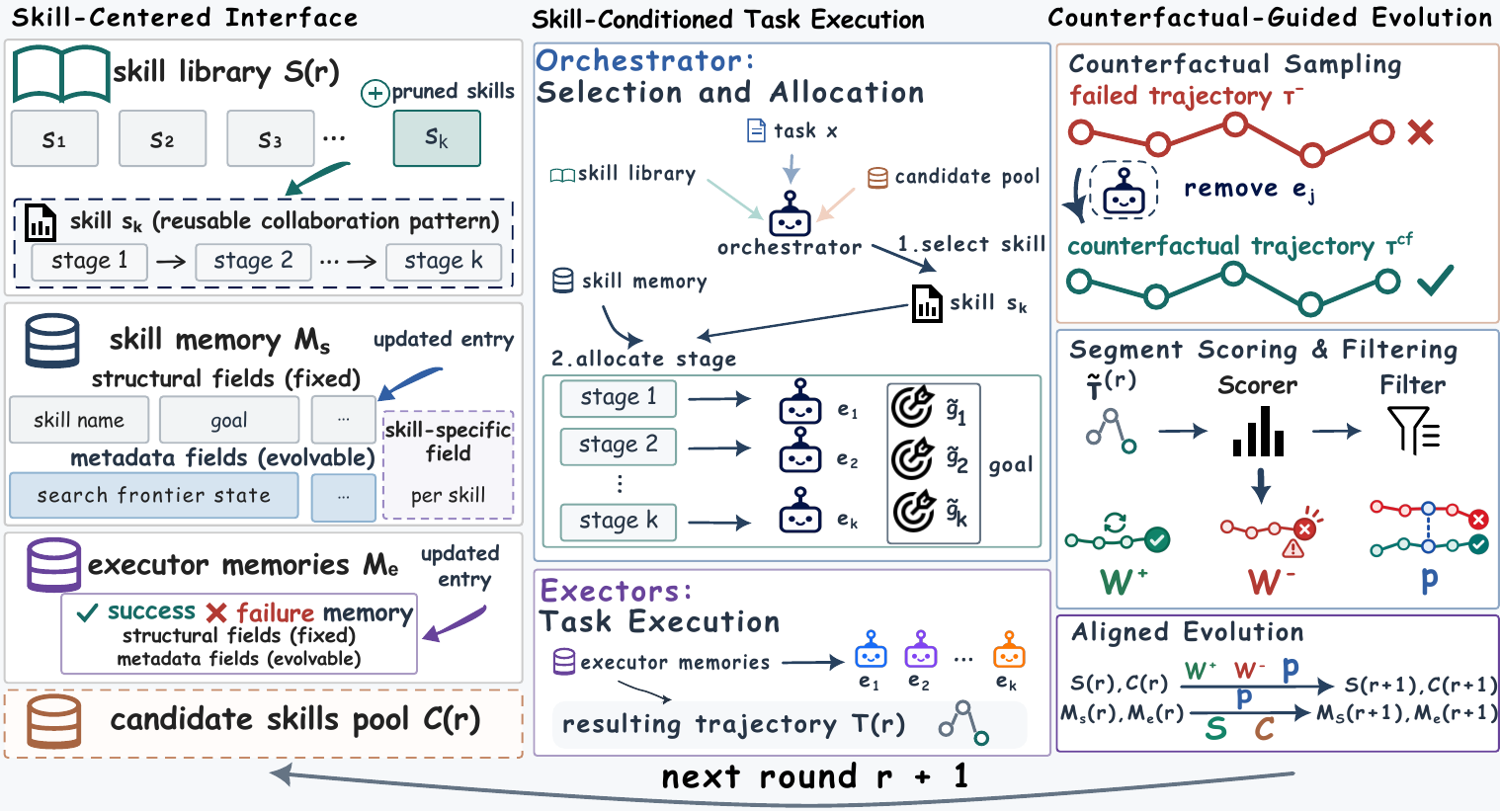}
    \caption{Overview of \mbox{CoEvolve}. Skills couple collaboration with supporting memories; skill-conditioned execution produces trajectories, and counterfactual comparisons yield local signals for aligned updates across rounds.}
    \label{fig:coevolve_overview}
\end{figure}

\subsection{Overview}
\label{sec:overview}

In this section, we detail the \mbox{CoEvolve} framework, as shown in Figure~\ref{fig:coevolve_overview}. We first introduce the skill-centered design, which aligns collaboration and memory to address cross-component mismatch. We then describe the skill-conditioned task execution process. Finally, we present counterfactual-guided evolution and explain how the resulting signals jointly update collaboration and memory.

\subsection{Skill-Centered Interface}
\label{sec:skill_interface}

\textbf{Skill.} To represent collaboration in a reusable and evolvable form, we model collaboration patterns as skills. As shown in Figure~\ref{fig:coevolve_overview}, a skill $s$ is defined by its purpose and applicable scenario. To support more complex collaboration, a skill may contain multiple stages, each of which is a finer-grained skill. Stages can be arranged in series or in parallel. New skills can be formed by composing existing skills or created directly.

\textbf{Memory.} To connect evolving collaboration patterns with the supporting experience, we organize memory around the skill space using skill memory $M_s$ and executor memory $M_e$. In $M_s$, structural fields identify the skill and context, while evolvable metadata fields specify which experience to store; a skill-specific field separates experience for different collaboration decisions. $M_e$ stores stage-level execution experience in success and failure memories. Both memories evolve with the skill space, allowing their support to adapt as collaboration patterns change.

\subsection{Skill-Conditioned Task Execution}
\label{sec:pipeline}

To operationalize the skill-centered interface during task execution, we organize each round as a skill-conditioned process. The orchestrator selects collaboration patterns and uses skill memory to allocate stages, while executors carry out assigned stages using their execution memories. We describe the two processes in turn.

\subsubsection{Orchestrator: Selection and Allocation}
\label{sec:selection}

At each decision step, the orchestrator selects a skill from the skill library $\mathcal{S}^{(r)}$ or the candidate pool $\mathcal{C}^{(r)}$ according to the current task context and specifies a corresponding goal $g_t$ for the selected skill. It then retrieves relevant experience from the skill memory $M_s^{(r)}$ using the retrieval mechanism described in Appendix~\ref{app:memory_retrieval}, and combines the retrieved experience with executor profiles to assign an executor and a stage-level goal to each stage. The entire decision process is written as:
\begin{equation}
(s_t, g_t, \pi_t) = \Phi\!\left(q_t, \mathcal{S}^{(r)}, \mathcal{C}^{(r)}, \mathrm{Ret}(q_t, M_s^{(r)}), \mathcal{E}\right),
\qquad
\pi_t = \{(e_k, \tilde{g}_k)\}_k.
\end{equation}
where $q_t$ is the current context, $\mathcal{C}^{(r)}$ is the candidate pool of skills generated in the previous round, $\mathrm{Ret}$ is the retriever used when allocating stages, $\mathcal{E}$ collects executors and their profiles, and $\pi_t$ assigns to each stage of the selected skill an executor $e_k$ and a stage-level goal $\tilde{g}_k$.

\subsubsection{Executors: Task Execution}
\label{sec:execution}
Each executor carries out its assigned stages by retrieving relevant experience from its success and failure memories $M_e^{(r)}$ and following the corresponding stage-level goal. For an executor $e_i$ assigned to stage $s$, its execution output is generated as
\begin{equation}
y_t = \mathrm{Exec}_{e_i}\left(o_t, g_{t,s}, \mathrm{Retrieve}\left(M_i^{(r)}, o_t, g_{t,s}\right)\right),
\end{equation}
where $o_t$ denotes the observation, $g_{t,s}$ is the goal assigned to stage $s$, and $y_t$ denotes the executor output, which may correspond to reasoning or an environment action. The environment returns feedback after each step, and the selection--allocation--execution cycle repeats until the task terminates, yielding the trajectory $\mathcal{T}^{(r)}$ together with the per-step records used by the evolution stage.

The resulting execution keeps collaboration and memory aligned: the selected skill defines the stages, skill memory informs their allocation, and executor memory supports stage-level execution.

\subsection{Counterfactual-Guided Evolution}
\label{sec:counterfactual}

After each round, observed trajectories provide only coarse feedback on which local decisions should be updated, requiring more localized evolution signals. \mbox{CoEvolve} therefore compares failed trajectories with counterfactual variants under the same context to derive such signals for aligned evolution of collaboration and memory. The process has three steps: counterfactual sampling, segment scoring and filtering, and aligned evolution.

\subsubsection{Counterfactual Sampling}
\label{sec:sampling}

Given a failed trajectory $\tau^{-}$, \mbox{CoEvolve} constructs a counterfactual trajectory by re-running the task from a branch point with one executor removed. Specifically, starting from a branch point $t_0$, the system preserves the original prefix of $\tau^{-}$ up to $t_0$ and re-executes the remaining process after removing executor $e_j$ from the available executors:
\begin{equation}
\tau_{\mathrm{cf}} = \mathrm{Sample}\left(\tau^{-},t_0,e_j\right).
\end{equation}
The generated counterfactual trajectory is then compared with its source trajectory based on the final outcome or average reward. If it succeeds or achieves a clear improvement, it is treated as an improved counterfactual trajectory; otherwise, it is treated as a non-improved one. Together with the original successful and failed trajectories, these counterfactual trajectories form the mining pool:
\begin{equation}
\widetilde{\mathcal{T}}^{(r)} = \mathcal{T}_{+}^{(r)} \cup \mathcal{T}_{-}^{(r)} \cup \mathcal{T}_{\mathrm{cf},+}^{(r)} \cup \mathcal{T}_{\mathrm{cf},-}^{(r)}.
\end{equation}
\subsubsection{Segment Scoring and Filtering}
\label{sec:extraction}

Given the mining pool $\widetilde{\mathcal{T}}^{(r)}$, \mbox{CoEvolve} first identifies local trajectory segments that provide informative signals for evolution. It enumerates short candidate segments $b$, each consisting of a local sequence of interaction steps or stage-level executions, and assigns each segment a score:
\begin{equation}
q(b) = \mathrm{Scorer}\left(b;\widetilde{\mathcal{T}}^{(r)}\right).
\end{equation}
The score reflects the informativeness of a segment based on step-level feedback, process rewards, novelty, repeated feedback, and its position in the trajectory. The scored segments are then filtered into three evidence sets:
\begin{equation}
(W_{+}^{(r)},W_{-}^{(r)},P^{(r)}) = \mathrm{Filter}\left(\widetilde{\mathcal{T}}^{(r)},q\right).
\end{equation}
Here, $W_{+}^{(r)}$ and $W_{-}^{(r)}$ collect useful and failure signals from successful/improved and failed/non-improved trajectories, respectively, with $W^{(r)}=W_{+}^{(r)}\cup W_{-}^{(r)}$. $P^{(r)}$ stores failed--improved counterfactual pairs, providing localized evidence around the branch point. The abstract formulation is provided in Appendix~\ref{app:segment_scoring}.

\subsubsection{Aligned Evolution of Collaboration and Memory}
\label{sec:evolution}

The filtered evidence is then used to jointly update the skill library and both memories.

\textbf{Skill evolution.} At round $r$, the candidate pool $\mathcal{C}^{(r)}$ has already participated in task execution together with the skill library $\mathcal{S}^{(r)}$. \mbox{CoEvolve} first uses the usage feedback $F^{(r)}$ collected during the round to promote useful candidates into the skill library and prune ineffective ones:
\begin{equation}
\mathcal{S}^{(r+1)} = \mathrm{Evolve}_{\mathrm{skill}}\left(\mathcal{S}^{(r)}, \mathcal{C}^{(r)}, F^{(r)}\right),
\end{equation}
and then uses the evolution signals $W^{(r)}$ to generate new candidate skills:
\begin{equation}
\mathcal{C}^{(r+1)} = \mathrm{Gen}_{\mathrm{cand}}\left(\mathcal{C}^{(r)}, W^{(r)}\right),
\end{equation}
which enter the candidate pool and are evaluated in the next round.

\textbf{Skill memory evolution.} Skill memory provides the experience needed for future stage allocation under the selected skills. \mbox{CoEvolve} therefore evolves its metadata fields using pivot samples in the context of the updated skill library and candidate pool, allowing the memory structure to adapt what information should be recorded for subsequent stage allocation:
\begin{equation}
M_s^{(r+1)} = \mathrm{Evolve}_{\mathrm{smem}}\left(M_s^{(r)}, \mathcal{S}^{(r+1)}, \mathcal{C}^{(r+1)}, P^{(r)}\right).
\end{equation}
Using $\mathcal{S}^{(r+1)}$ and $\mathcal{C}^{(r+1)}$ during memory evolution allows the memory structure to adapt to changes in both established and candidate skills.

\textbf{Executor memory evolution.} Executor memories capture stage-level execution experience. For each executor, \mbox{CoEvolve} selects pivot samples in which removing that executor leads to an improved outcome. These agent-specific contrasts, interpreted in the context of the updated skill library and candidate pool, guide the evolution of the metadata fields in its success and failure memories:
\begin{equation}
M_e^{(r+1)} = \mathrm{Evolve}_{\mathrm{emem}}\left(M_e^{(r)}, \mathcal{S}^{(r+1)}, \mathcal{C}^{(r+1)}, P^{(r)}\right),
\end{equation}
where $M_e^{(r)}=\{M_i^{(r)}\}_{i=1}^{N}$ collects the memories of all executors. This changes what stage-level experience can be recorded in future entries, which are populated during execution.

In this way, \mbox{CoEvolve} uses counterfactual sampling to test local alternatives under the same context, rather than relying solely on task-level scores or post-hoc analysis of observed trajectories. This allows it to identify outcome-sensitive local signals that are directly informative for evolution. These signals then jointly drive updates to collaboration patterns and their supporting memories, keeping the two components coordinated across rounds.

\section{Experiments}
\label{sec:experiments}

In this section, we first describe the experimental setup, including the datasets, evaluation metrics, and baselines, and then present the experimental results.

\subsection{Experimental Setup}
\label{sec:experimental_setup}

\textbf{Datasets.}
We evaluate \mbox{CoEvolve} on two widely used interactive decision-making benchmarks: ALFWorld~\citep{shridhar2020alfworld} and WebShop~\citep{yao2022webshop}. ALFWorld is a text-based embodied environment for household tasks, and we use its valid-unseen split of 134 tasks. WebShop simulates web-based shopping interactions, for which we use the official test set of 500 tasks. Further details are provided in Appendix~\ref{app:benchmarks}.

\textbf{Evaluation Metrics.}
We report task success rate (SR) on ALFWorld and the purchased-product score on WebShop. We additionally report the average execution length on both benchmarks.

\textbf{Baselines.}
We compare \mbox{CoEvolve} with representative collaboration- and memory-based methods. Collaboration-based baselines include DyLAN~\citep{liu2024dylan} and MacNet~\citep{qian2025macnet}, while memory-based baselines include ExpeL~\citep{zhao2024expel}, AWM~\citep{wang2025awm}, EvolveR~\citep{wu2025evolver}, MemEvolve~\citep{zhang2025memevolve}, and AutoGen-GMemory~\citep{wu2023autogen,zhang2025gmemory}. Detailed descriptions of the baselines are provided in Appendix~\ref{app:baseline_details}. Unless otherwise specified, all methods are evaluated using Qwen3-235B-A22B-Instruct-2507~\citep{qwen2025qwen3235b2507}; full implementation details and prompts are provided in Appendices~\ref{app:implementation} and~\ref{app:prompts}.

\begin{table}[!ht]
\caption{Performance on ALFWorld and WebShop. ALFWorld columns report overall and task-wise success rates; WebShop reports purchased-product score.}
\label{tab:main_results}
\centering
\normalsize
\setlength{\tabcolsep}{0pt}
\renewcommand{\arraystretch}{1.15}
\newcommand{\resultrow}[9]{%
  \makebox[93pt][l]{#1} & \makebox[40pt][c]{#2} &
  \makebox[36pt][c]{#3} & \makebox[36pt][c]{#4} &
  \makebox[36pt][c]{#5} & \makebox[36pt][c]{#6} &
  \makebox[36pt][c]{#7} & \makebox[36pt][c]{#8} &
  \makebox[40pt][c]{#9}\\}
\begin{tabular}{@{}lcccccccc@{}}
\toprule
\makebox[93pt][l]{Dataset} & \multicolumn{7}{c}{ALFWorld} & \makebox[40pt][c]{WebShop}\\
\midrule
\resultrow{Method}{Overall}{put}{clean}{heat}{cool}{examine}{puttwo}{Score}
\midrule
\resultrow{DyLAN}{$0.6940$}{$0.7500$}{$0.6129$}{$0.6522$}{$0.6667$}{$0.8333$}{$0.7059$}{$0.5935$}
\resultrow{MemEvolve}{$0.7239$}{$0.8333$}{$0.6452$}{$0.7826$}{$0.6190$}{$0.8333$}{$0.6471$}{$0.5249$}
\resultrow{EvolveR}{$0.7015$}{$0.7917$}{$0.6774$}{$0.8261$}{$0.7619$}{$0.6111$}{$0.4706$}{$0.5715$}
\resultrow{AWM}{$0.6716$}{$\mathbf{0.8750}$}{$0.7097$}{$0.5652$}{$0.6190$}{$0.6667$}{$0.5294$}{$0.5364$}
\resultrow{AutoGen-GMemory}{$0.7537$}{$0.8333$}{$0.7097$}{$\mathbf{0.8696}$}{$\mathbf{0.8095}$}{$0.7778$}{$0.4706$}{$0.6681$}
\resultrow{MacNet}{$0.7090$}{$0.7917$}{$0.6452$}{$0.7391$}{$\mathbf{0.8095}$}{$0.7222$}{$0.5294$}{$0.6450$}
\resultrow{ExpeL}{$0.7463$}{$\mathbf{0.8750}$}{$\mathbf{0.7419}$}{$0.6522$}{$0.7143$}{$0.7778$}{$0.7059$}{$0.5739$}
\midrule
\multicolumn{9}{@{}l@{}}{%
  \begingroup\setlength{\fboxsep}{0pt}%
  \colorbox{black!8}{%
    \begin{tabular}{@{}lcccccccc@{}}%
    \resultrow{\textbf{CoEvolve}}{$\mathbf{0.7985}$}{$\mathbf{0.8750}$}{$0.6774$}{$0.7826$}{$\mathbf{0.8095}$}{$\mathbf{0.8889}$}{$\mathbf{0.8235}$}{$\mathbf{0.7072}$}
    \end{tabular}%
  }%
  \endgroup%
}\\
\bottomrule
\end{tabular}
\end{table}

\subsection{Overall Performance}
\label{sec:overall_performance}

To demonstrate the effectiveness of \mbox{CoEvolve}, we compare it with representative collaboration- and memory-based baselines, and the results are shown in Table~\ref{tab:main_results}. From the results, we can make the following observations:

\begin{enumerate}
\setlength{\topsep}{2pt}
\setlength{\itemsep}{0pt}
\setlength{\parsep}{0pt}
\item[(1)] \mbox{CoEvolve} achieves the best overall performance on both benchmarks. On ALFWorld, \mbox{CoEvolve} achieves a success rate of $0.7985$, outperforming the strongest baseline, AutoGen-GMemory ($0.7537$), by $4.48$ percentage points. On WebShop, \mbox{CoEvolve} achieves a score of $0.7072$, exceeding AutoGen-GMemory ($0.6681$) by $3.91$ points on a 100-point scale. The consistent improvements across two distinct interactive environments demonstrate the effectiveness of jointly evolving collaboration and memory.
\item[(2)] The task-wise results on ALFWorld show that the overall improvement is not driven by a single task type. \mbox{CoEvolve} achieves or matches the best performance on four of the six task types, including \emph{put}, \emph{cool}, \emph{examine}, and \emph{puttwo}. In particular, it achieves $0.8889$ on \emph{examine} and $0.8235$ on \emph{puttwo}, outperforming the strongest baselines by $5.56$ and $11.76$ percentage points, respectively. This indicates that \mbox{CoEvolve} can effectively adapt its collaboration patterns and supporting experience to different task requirements.
\item[(3)] Existing collaboration- and memory-based methods exhibit varying strengths across different task types. For example, AutoGen-GMemory performs best on \emph{heat}, while ExpeL achieves the strongest baseline performance on \emph{clean} and \emph{puttwo}. In contrast, \mbox{CoEvolve} achieves the highest overall performance on both benchmarks while maintaining competitive performance across different task types, rather than excelling only on specific ones. These results suggest that jointly adapting collaboration and memory provides more consistent improvements than evolving either aspect independently.

\end{enumerate}

\subsection{Effect of Aligned Co-Evolution}
\label{sec:aligned_coevolution}

\begin{table}[!ht]
\caption{Ablation studies on ALFWorld and WebShop. Blocks compare (a) evolved components and (b) counterfactual sampling and delayed promotion; stages are average executed skill stages.}
\label{tab:ablation}
\centering
\normalsize
\renewcommand{\arraystretch}{1.08}
\setlength{\tabcolsep}{7pt}
\begin{tabular}{@{}lcccc@{}}
\toprule
\multicolumn{1}{c}{} & \multicolumn{2}{c}{ALFWorld} & \multicolumn{2}{c}{WebShop} \\
\cmidrule(lr){2-3}\cmidrule(l){4-5}
Setting & SR $\uparrow$ & Stages $\downarrow$ & Score $\uparrow$ & Stages $\downarrow$ \\
\midrule
\multicolumn{5}{@{}l@{}}{\textbf{(a) Evolved components}} \\
\addlinespace[2pt]
None & $0.6119$ & $21.4328$ & $0.5238$ & $16.7540$ \\
Skill & $0.6567$ & $20.1496$ & $0.5718$ & $15.2360$ \\
SMem & $0.6418$ & $21.0149$ & $0.5521$ & $15.6730$ \\
EMem & $0.6194$ & $20.7090$ & $0.5763$ & $16.1980$ \\
SMem + EMem & $0.6940$ & $19.2388$ & $0.5996$ & $14.6850$ \\
Skill + EMem & $0.6791$ & $18.7910$ & $0.6404$ & $13.5370$ \\
Skill + SMem & $0.7015$ & $18.4254$ & $0.6293$ & $13.1460$ \\
\midrule
\rowcolor{black!8}
\textbf{Skill + SMem + EMem (CoEvolve)} & $\mathbf{0.7985}$ & $\mathbf{16.8806}$ & $\mathbf{0.7072}$ & $\mathbf{11.7520}$ \\
\specialrule{0.8pt}{5pt}{5pt}
\multicolumn{5}{@{}l@{}}{\textbf{(b) Counterfactual sampling and promotion}} \\
\addlinespace[2pt]
w/o Counterfactual & $0.6791$ & $19.7537$ & $0.5768$ & $15.2980$ \\
w/o Promotion & $0.7164$ & $17.3881$ & $0.6395$ & $13.6060$ \\
\midrule
\rowcolor{black!8}
\textbf{CoEvolve} & $\mathbf{0.7985}$ & $\mathbf{16.8806}$ & $\mathbf{0.7072}$ & $\mathbf{11.7520}$ \\
\bottomrule
\end{tabular}
\end{table}

To examine whether collaboration and memory benefit from aligned evolution, we vary the components allowed to evolve while keeping the remaining components fixed. The results are shown in Table~\ref{tab:ablation}(a). \textbf{(1)} Evolving individual components generally improves performance over the non-evolving setting, while combining Skill with memory brings further gains. In particular, Skill + SMem and Skill + EMem outperform their corresponding single-component variants, showing that evolving collaboration benefits from jointly adapting its supporting experience. Further evolving both SMem and EMem with Skill achieves $0.7985$ on ALFWorld and $0.7072$ on WebShop, outperforming the strongest partial configurations by $9.70$ and $6.68$ percentage points, respectively.

\noindent\begin{minipage}{\linewidth}
This demonstrates the benefit of aligning evolving collaboration patterns with both collaboration- and execution-level experience. \textbf{(2)} Such aligned co-evolution also improves execution efficiency: the full model reduces the average execution stages from $21.43$ to $16.88$ on ALFWorld and from $16.75$ to $11.75$ on WebShop, and requires fewer stages than all single- and dual-component variants.
\end{minipage}

\subsection{Counterfactual-Guided Evolution}
\label{sec:counterfactual_guided_ablation}

We further examine counterfactual sampling for guiding evolution and delayed promotion for validating candidate skills. Table~\ref{tab:ablation}(b) reports the results. \textbf{(1)} Removing counterfactual sampling reduces performance from $0.7985$ to $0.6791$ on ALFWorld and from $0.7072$ to $0.5768$ on WebShop, showing that contrasting failed trajectories with improved counterfactual continuations provides more informative signals for identifying effective evolution directions. \textbf{(2)} Removing delayed promotion reduces performance to $0.7164$ and $0.6395$, while increasing the average execution stages from $16.88$ to $17.39$ and from $11.75$ to $13.61$ on ALFWorld and WebShop, respectively. This suggests that evaluating candidate skills before promotion helps filter ineffective updates and prevents them from being directly incorporated into the skill library.

\subsection{Evolution Behavior Analysis}
\label{sec:evolution_behavior_analysis}

\begin{figure}[!b]
\centering
\begin{minipage}[t]{0.49\linewidth}
\centering
\includegraphics[width=\linewidth]{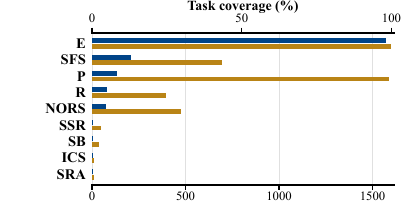}\\
\textbf{(a)}
\end{minipage}\hfill%
\begin{minipage}[t]{0.49\linewidth}
\centering
\includegraphics[width=\linewidth]{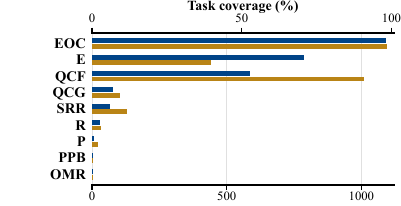}\\
\textbf{(b)}
\end{minipage}
\caption{Skill calls and task coverage: (a) ALFWorld; (b) WebShop.}
\label{fig:skill_usage}
\end{figure}

To understand how the collaboration structure evolves in practice, we analyze the call frequency and task coverage of learned skills on ALFWorld and WebShop (Figure~\ref{fig:skill_usage}). On ALFWorld, original skills remain the backbone of task execution, while evolved skills are repeatedly invoked across different tasks. On WebShop, several evolved skills likewise show substantial use and coverage. These patterns indicate that \mbox{CoEvolve} learns reusable collaboration patterns rather than trajectory-specific modifications. Other evolved skills have lower coverage, consistent with more specialized behaviors for conflict or recovery states. Overall, the evolved skill library contains both broadly reusable routines and more targeted behaviors.

\subsection{Robustness Analysis}
\label{sec:robustness_analysis}
\label{sec:orchestrator_capacity}

\textbf{Orchestrator capacity.} To assess whether the effectiveness of \mbox{CoEvolve} depends on orchestrator capacity, we compare Qwen3.5-Plus~\citep{qwen2026qwen35plus} with the weaker Qwen3.5-35B-A3B~\citep{qwen2026qwen3535ba3b} in Figure~\ref{fig:orchestrator_capacity}. Collaboration-only evolves Skill, while memory-only evolves SMem and EMem. \mbox{CoEvolve} consistently achieves the largest gains on both benchmarks under both orchestrators, showing that the benefit of joint evolution persists across different orchestrator capacities. The smaller gains with Qwen3.5-35B-A3B further suggest that orchestrator capacity affects the quality of evolution, but does not eliminate the advantage of jointly evolving collaboration and memory across different settings.

\subsection{Execution Efficiency}
\label{sec:execution_efficiency}

We additionally compare the average execution length of CoEvolve and the baselines in Figure~5. CoEvolve requires only 16.88 execution stages on ALFWorld and 11.75 on WebShop, achieving shorter execution than the baselines on both benchmarks. Combined with the performance gains in Table~1, this suggests that the evolved collaboration structure enables more effective task solving with fewer execution steps.

\par\noindent
\begin{minipage}[t]{0.49\linewidth}
\vspace{0pt}
\centering
\includegraphics[width=\linewidth,height=0.48\linewidth,keepaspectratio,trim=0 15pt 0 3pt,clip]{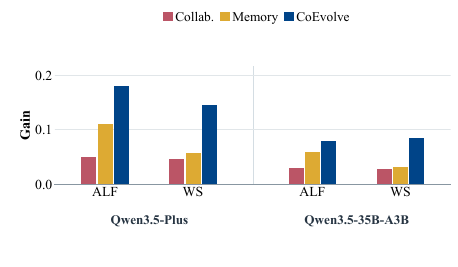}
\captionof{figure}{Performance gains over no evolution under different orchestrator capacities.}
\label{fig:orchestrator_capacity}
\end{minipage}\hfill
\begin{minipage}[t]{0.49\linewidth}
\vspace{0pt}
\centering
\includegraphics[width=\linewidth,height=0.48\linewidth,keepaspectratio,trim=0 1pt 0 12pt,clip]{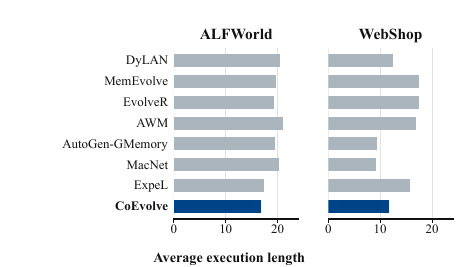}
\captionof{figure}{Average execution length on ALFWorld and WebShop.}
\label{fig:execution_length}
\end{minipage}
\par

\section{Conclusion}
\label{sec:conclusion}

In this paper, we introduce \mbox{CoEvolve} as a framework for the aligned co-evolution of collaboration and memory in LLM-based multi-agent systems. It represents reusable collaboration patterns through a shared skill space, which provides a common interface for evolving collaboration decisions and their supporting memories. The framework further uses counterfactual trajectory comparisons to identify outcome-sensitive local signals and guide coordinated updates to the skill library, skill memory, and executor memories. Experiments on ALFWorld and WebShop show that \mbox{CoEvolve} achieves the best overall performance among the evaluated methods. Ablation studies confirm that jointly evolving collaboration and memory provides greater gains than evolving either side in isolation, while counterfactual-guided evolution supplies more informative signals for system improvement. Robustness analyses show that these benefits remain consistent across different orchestrator capacities. Overall, \mbox{CoEvolve} provides a unified approach for evolving how multi-agent systems collaborate together with the experience that supports their collaboration.

\section*{AI Use Statement}

In this work, the authors used generative AI tools for code completion, debugging suggestions, LaTeX formatting, language polishing, and experimental result analysis. The authors carefully reviewed all AI-assisted work and independently verified all research ideas, experiments, analyses, and final content. The authors take responsibility for the final content of this work, including text, claims, and artifacts produced with the aid of generative AI.

\section*{Ethics Statement}

This work studies the self-evolution of LLM-based multi-agent systems in simulated interactive environments. The experiments use publicly available benchmarks and do not involve human participants, private personal data, or real-world deployment. The proposed method is intended for research on agent collaboration and adaptation; potential risks associated with deploying autonomous multi-agent systems in real-world settings are outside the scope of the current evaluation.

\section*{Reproducibility Statement}

The method and evaluation protocol are described in Sections~\ref{sec:method} and~\ref{sec:experiments}. Appendices~\ref{app:memory_retrieval}--\ref{app:implementation} provide retrieval and scoring formulations, benchmark and baseline descriptions, and implementation details; Appendices~\ref{app:additional_exp}--\ref{app:prompts} contain additional analyses, executor descriptions, case studies, and prompt templates. The anonymized code is available at \url{https://anonymous.4open.science/r/Co-Evolve-8239}.

\bibliography{iclr2027_conference}
\bibliographystyle{iclr2027_conference}

\clearpage
\appendix
\section{Memory Retrieval}
\label{app:memory_retrieval}

For both skill memory and executor memories, \mbox{CoEvolve} uses a unified retrieval function inspired by SimUtil-UCB~\citep{zhang2026retroagent}. It scores an item $m$ in memory set $M$ for query $q$ using semantic similarity, historical utility, and an exploration bonus:
\begin{equation}
\rho(q,m)=\lambda_1s(q,m)+\lambda_2u(m)+\lambda_3e(m).
\label{eq:memory_score}
\end{equation}
Here $s(q,m)$ is semantic similarity, $u(m)$ is historical utility, and $e(m)$ is the exploration bonus. Retrieval selects the highest-scoring items:
\begin{equation}
\operatorname{Ret}(q,M)=\operatorname{TopK}_{m\in M}\rho(q,m).
\label{eq:memory_retrieval}
\end{equation}
The three coefficients control the relative weights of these terms. All reported experiments use $\lambda_1=\lambda_2=\lambda_3=1$.

\section{Segment Scoring and Pivot Identification}
\label{app:segment_scoring}

\subsection{Step-Level Scoring}
For a trajectory $\tau=(z_1,\ldots,z_T)$, a step scorer assigns
\begin{equation}
u_t=\operatorname{StepScore}_{\psi}(z_t,r_t,f_{\leq t}),
\label{eq:step_score}
\end{equation}
where $z_t$ is a structured interaction step, $r_t$ is process reward or feedback, and $f_{\leq t}$ is meaningful feedback observed through step $t$. The score summarizes task progress, feedback diversity, and repeated feedback.

\subsection{Segment-Level Scoring}
For a candidate segment $b_{i:j}=(z_i,\ldots,z_j)$, define its context as
\begin{equation}
c_{i:j}=\bigl(\operatorname{pos}(b_{i:j}),\operatorname{type}(\tau)\bigr),
\label{eq:segment_context}
\end{equation}
where position records where the segment occurs and trajectory type distinguishes original successes, original failures, improved counterfactuals, and non-improved counterfactuals. The segment scorer then aggregates local evidence:
\begin{equation}
q(b_{i:j})=\operatorname{SegScore}_{\psi}\bigl(b_{i:j},\{u_t\}_{t=i}^{j},c_{i:j}\bigr).
\label{eq:segment_aggregation}
\end{equation}
This score is used to filter useful and harmful local patterns into $W_{+}^{(r)}$ and $W_{-}^{(r)}$, respectively.

\subsection{Pivot Identification}
For a failed trajectory and an improved counterfactual trajectory branching at $t_0$, define the cumulative step-score difference
\begin{equation}
\Delta_t=\sum_{k=t_0}^{t}\bigl(u_k^{\mathrm{cf}}-u_k^{-}\bigr),
\label{eq:pivot_difference}
\end{equation}
and select the first positive crossing as
\begin{equation}
t^*=\min\{t\mid\Delta_t>0\}.
\label{eq:pivot_search}
\end{equation}
If no positive crossing exists, this rule does not yield a pivot. When a pivot exists, the paired segments form a localized comparison sample in $P^{(r)}$.

\section{Benchmark Details}
\label{app:benchmarks}

\paragraph{ALFWorld.} ALFWorld is a text-based embodied decision-making benchmark that aligns interactive text environments with household tasks from ALFRED~\citep{shridhar2020alfworld}. Agents observe textual environment descriptions and issue text actions to navigate, find, and manipulate objects. We evaluate on the valid-unseen split of 134 tasks using task success rate, with at most 30 interaction steps per task.

\paragraph{WebShop.} WebShop is a simulated e-commerce website benchmark for grounded web interaction~\citep{yao2022webshop}. Each task specifies a shopping goal; the agent searches for items, inspects product pages, selects options, and attempts a purchase. We use the official 500-task test set and report purchased-product score, with at most 30 interaction steps per task.

\section{Baseline Details}
\label{app:baseline_details}

We provide detailed descriptions of the baseline methods used in our experiments.

\paragraph{Collaboration-based methods.}
We include the following collaboration-based baselines:
\begin{itemize}
    \raggedright
    \setlength{\itemsep}{0pt}
    \setlength{\topsep}{2pt}

    \item \textbf{DyLAN}~\citep{liu2024dylan} dynamically organizes agent interactions according to task requirements, enabling task-specific collaboration among agents.

    \item \textbf{MacNet}~\citep{qian2025macnet} coordinates multiple agents through a structured collaboration network that determines how information and intermediate results are exchanged during task execution.
\end{itemize}

\paragraph{Memory-based methods.}
We include the following memory-based baselines:
\begin{itemize}
    \raggedright
    \setlength{\itemsep}{0pt}
    \setlength{\topsep}{2pt}

    \item \textbf{ExpeL}~\citep{zhao2024expel} extracts reusable insights from past experience and applies them to improve performance on future tasks.

    \item \textbf{AWM}~\citep{wang2025awm} induces reusable workflows from prior experience and retrieves relevant workflows to guide subsequent task execution.

    \item \textbf{EvolveR}~\citep{wu2025evolver} iteratively refines reusable experience based on feedback collected during interaction and task solving.

    \item \textbf{MemEvolve}~\citep{zhang2025memevolve} jointly evolves experiential knowledge and the underlying memory architecture, allowing both stored experience and memory organization to adapt over time.

    \item \textbf{AutoGen-GMemory}~\citep{wu2023autogen,zhang2025gmemory} equips multi-agent systems with structured collaboration memory, allowing agents to retain and reuse information from previous interactions.
\end{itemize}

\section{Implementation Details}
\label{app:implementation}

\subsection{Evolution Configuration}
We run \mbox{CoEvolve} for three evolution rounds, sampling 50 tasks from the official training split in each round. No validation/test tasks are used during evolution. Counterfactual trajectories are constructed from at most 20 failed tasks per round. In each round, we generate four candidate skills: two from successful trajectories, one from improved counterfactuals, and one from failure cases. These candidates are evaluated in the following round before being promoted to the skill library. We use Qwen3.5-Plus~\citep{qwen2026qwen35plus} for the orchestrator and evolution modules, and GPT-4.1-nano~\citep{openai2025gpt41nano} for the executors during evolution. During evaluation, we use Qwen3-235B-A22B-Instruct-2507~\citep{qwen2025qwen3235b2507} for both the orchestrator and executors of \mbox{CoEvolve}, as well as for all baselines. For memory retrieval, we use all-MiniLM-L6-v2 as the embedding model and set $\lambda_1=\lambda_2=\lambda_3=1$.

\newpage
\subsection{Complete Workflow}
Algorithm~\ref{alg:coevolve} collects the operations defined in Section~\ref{sec:counterfactual} into one round-level procedure. The candidate pool generated at the end of round $r$ is tried in round $r+1$ before any promotion decision.

\begin{algorithm}[H]
\caption{CoEvolve workflow across evolution rounds}
\label{alg:coevolve}
\begin{algorithmic}[1]
\Require Initial $\mathcal{S}^{(0)},\mathcal{C}^{(0)},M_s^{(0)},M_e^{(0)}$; task batches $\{\mathcal{B}^{(r)}\}_{r=0}^{R-1}$
\For{$r=0,\ldots,R-1$}
  \State Execute each $x\in\mathcal{B}^{(r)}$ under $\mathcal{S}^{(r)}\cup\mathcal{C}^{(r)}$ using $M_s^{(r)}$ and $M_e^{(r)}$.
  \State Collect trajectories $\mathcal{T}^{(r)}$ and candidate-usage feedback $F^{(r)}$.
  \State Sample counterfactuals from at most 20 failed tasks: $\mathcal{T}_{\mathrm{cf}}^{(r)}\gets\mathrm{CF}(\mathcal{T}_{-}^{(r)})$.
  \State $\widetilde{\mathcal{T}}^{(r)}\gets\mathcal{T}^{(r)}\cup\mathcal{T}_{\mathrm{cf}}^{(r)}$.
  \State $(W_{+}^{(r)},W_{-}^{(r)},P^{(r)})\gets\mathrm{Filter}(\widetilde{\mathcal{T}}^{(r)},q)$; $W^{(r)}\gets W_{+}^{(r)}\cup W_{-}^{(r)}$.
  \State $\mathcal{S}^{(r+1)}\gets\mathrm{Evolve}_{\mathrm{skill}}(\mathcal{S}^{(r)},\mathcal{C}^{(r)},F^{(r)})$.
  \State $\mathcal{C}^{(r+1)}\gets\mathrm{Gen}_{\mathrm{cand}}(\mathcal{C}^{(r)},W^{(r)})$.
  \State $M_s^{(r+1)}\gets\mathrm{Evolve}_{\mathrm{smem}}(M_s^{(r)},\mathcal{S}^{(r+1)},\mathcal{C}^{(r+1)},P^{(r)})$.
  \State $M_e^{(r+1)}\gets\mathrm{Evolve}_{\mathrm{emem}}(M_e^{(r)},\mathcal{S}^{(r+1)},\mathcal{C}^{(r+1)},P^{(r)})$.
\EndFor
\State \Return $\mathcal{S}^{(R)},\mathcal{C}^{(R)},M_s^{(R)},M_e^{(R)}$.
\end{algorithmic}
\end{algorithm}

\section{Additional Experimental Results}
\label{app:additional_exp}

\subsection{Hyperparameter Sensitivity}
\label{app:hyperparameter_sensitivity}

Table~\ref{tab:hyperparameter} reports the sensitivity of \mbox{CoEvolve} to the weighting coefficients $\lambda_1$, $\lambda_2$, and $\lambda_3$. The default setting $(1,1,1)$ achieves the best performance on both benchmarks. Across alternative configurations, performance ranges from $0.7164$ to $0.7687$ on ALFWorld and from $0.6629$ to $0.6863$ on WebShop, indicating that \mbox{CoEvolve} remains reasonably robust to different weight configurations while balanced weighting performs best.

\begin{table}[H]
\caption{Retrieval-weight sensitivity on the evolved \mbox{CoEvolve} system.}
\label{tab:hyperparameter}
\centering
\normalsize
\setlength{\tabcolsep}{6pt}
\begin{tabular}{@{}lccccc@{}}
\toprule
Setting & Sem. $\lambda_1$ & Util. $\lambda_2$ & Explor. $\lambda_3$ & ALF. SR & WS Score\\
\midrule
Default & 1.0 & 1.0 & 1.0 & \textbf{0.7985} & \textbf{0.7072}\\
Semantic $\times 0.5$ & 0.5 & 1.0 & 1.0 & 0.7463 & 0.6732\\
Semantic $\times 1.5$ & 1.5 & 1.0 & 1.0 & 0.7687 & 0.6863\\
Utility $\times 0.5$ & 1.0 & 0.5 & 1.0 & 0.7388 & 0.6629\\
Utility $\times 1.5$ & 1.0 & 1.5 & 1.0 & 0.7612 & 0.6781\\
Exploration $\times 0.5$ & 1.0 & 1.0 & 0.5 & 0.7164 & 0.6748\\
Exploration $\times 1.5$ & 1.0 & 1.0 & 1.5 & 0.7388 & 0.6807\\
\bottomrule
\end{tabular}
\end{table}

\subsection{Evolution Overhead}
\label{app:overhead}
Table~\ref{tab:evolution_overhead} reports the ALFWorld counterfactual-sampling cost across three evolution rounds. The average per round is 3.912 million tokens and 5,541 seconds; this is offline evolution overhead, not task-time execution length.
\begin{table}[H]
\caption{Counterfactual sampling overhead on ALFWorld across three evolution rounds.}
\label{tab:evolution_overhead}
\centering
\begin{tabular}{lcc}
\toprule
Round & Tokens & Time (s)\\
\midrule
R1 & 4.286M & 6,090\\
R2 & 3.704M & 5,238\\
R3 & 3.746M & 5,295\\
\midrule
\textbf{Avg.} & \textbf{3.912M} & \textbf{5,541}\\
\bottomrule
\end{tabular}
\end{table}

\subsection{Executor Specialization Analysis}
\label{app:executor_specialization}
\begin{figure}[H]
\begin{minipage}[t]{0.49\linewidth}
\vspace{0pt}
\centering
\includegraphics[width=\linewidth]{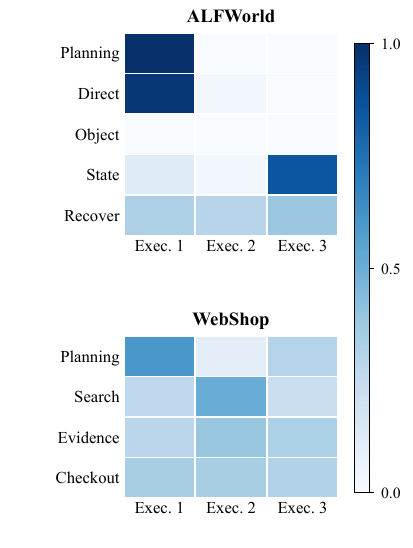}
\caption{Skill-level executor participation on ALFWorld and WebShop.}
\label{fig:executor_skill_categories}
\end{minipage}\hfill
\begin{minipage}[t]{0.49\linewidth}
\vspace{0pt}
\centering
\includegraphics[width=\linewidth]{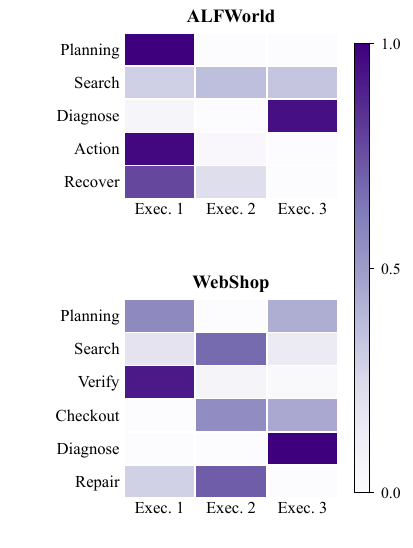}
\caption{Stage-level executor assignments on ALFWorld and WebShop.}
\label{fig:executor_stage_categories}
\end{minipage}
\end{figure}

We group skills and stages into functional categories and normalize executor assignments within each category. Figure~\ref{fig:executor_skill_categories} shows skill-level participation; Figure~\ref{fig:executor_stage_categories} shows the finer-grained stage assignments. These analyses characterize observed roles rather than establishing that memory alone caused specialization.

ALFWorld's planning and direct-action skill categories are concentrated on Executor 1, while its state category is concentrated on Executor 3. On WebShop, skill-level participation is less concentrated, but Executor 2 participates more in search and evidence categories. At the stage level, ALFWorld action stages concentrate on Executor 1 and diagnosis stages on Executor 3; WebShop checkout and repair stages favor Executor 2 while diagnosis favors Executor 3. These differences are consistent with a non-uniform division of work across executors.

\section{Executor Personas and Specialization}
\label{app:executor_personas}

Table~\ref{tab:executor_personas} reproduces the executor personas used in the earlier implementation. They describe intended tendencies rather than deterministic stage assignments.
\begin{table}[H]
\caption{Personas assigned to the three executors.}
\label{tab:executor_personas}
\centering
\begin{tabular}{@{}p{0.19\linewidth}p{0.73\linewidth}@{}}
\toprule
\textbf{Executor} & \textbf{Persona}\\
\midrule
Executor 1 & A calm and methodical thinker who brings structure to messy situations, surfaces implicit assumptions, and works carefully within defined boundaries.\\
Executor 2 & A steady and results-oriented contributor who turns direction into momentum, makes practical decisions, and keeps work moving without unnecessary overhead.\\
Executor 3 & A rigorous and independent-minded evaluator who pressure-tests ideas, notices fragile assumptions, and improves outcomes by identifying what may fail.\\
\bottomrule
\end{tabular}
\end{table}

The patterns in Appendix~\ref{app:executor_specialization} are broadly consistent with these descriptions: Executor 1 is often involved in planning and structured advancement, Executor 2 in search or concrete actions, and Executor 3 in diagnosis or failure-sensitive stages. This correspondence does not establish that the personas caused the observed allocation patterns.

\section{Trajectory-Level Case Study}
\label{app:case_study}

This appendix provides two trajectory excerpts illustrating the skill-usage behavior discussed in Section~\ref{sec:evolution_behavior_analysis}. The fields are retained from the earlier recorded cases; ellipses omit less relevant log fragments. These examples illustrate behavior, not aggregate effectiveness.

\begingroup
\tcbset{
    casebox/.style={
        breakable,
        enhanced,
        colback=black!2,
        colframe=black!25,
        coltitle=black,
        fonttitle=\bfseries,
        boxrule=0.25pt,
        arc=0pt,
        outer arc=0pt,
        left=4pt,
        right=4pt,
        top=4pt,
        bottom=4pt,
        boxsep=2pt,
        before skip=0.55\baselineskip,
        after skip=0.75\baselineskip,
        toptitle=2pt,
        bottomtitle=2pt
    }
}

\newcommand{\casestep}[1]{\medskip\noindent\textbf{#1}\par\smallskip}
\newcommand{\casefield}[2]{\noindent\textbf{#1.} #2\par\smallskip}
\newcommand{\caseellipsis}{\medskip\centerline{$\cdots$}\medskip}

\begin{tcolorbox}[casebox, title={Case 1: ALFWorld trajectory with common and evolved search skills}]
\casefield{Task}{Examine the book with the desklamp.}

\casestep{Task 102 step 1}

\casefield{Skill}{plan (participants=['Executor\_1'])}

\casefield{Executor output [Executor\_1/plan\_stage]}{To examine the book with the desklamp, I must first locate both the book and the desklamp. ... I will start by navigating to desk 1 ... This step is purely planning: no action is required yet.}

\casefield{Environment Feedback}{No environment action in this step}

\casestep{Task 102 step 2}

\casefield{Skill}{execute (participants=['Executor\_1'])}

\casefield{Executor output [Executor\_1/execute\_stage]}{Following the skill-level subgoal, I must go to desk 1 first to verify the co-location of both the book and desklamp. ...}

\casefield{Executor action [Executor\_1]}{go to desk 1}

\casefield{Environment Feedback}{You arrive at desk 1. On the desk 1, you see a alarmclock 1, a bowl 2, a cd 1, a cellphone 3, a desklamp 1, a mug 1, a pen 1, and a pencil 1.}

\caseellipsis

\casestep{Task 102 step 4}

\casefield{Skill}{reflect (participants=['Executor\_1'])}

\casefield{Executor output [Executor\_1/reflect\_stage]}{The current trace confirms that the book was not found on desk 1 or desk 2, while the desklamp is present only on desk 1. ... the next action class must be a systematic search of high-probability receptacles ...}

\casefield{Environment Feedback}{No environment action in this step}

\caseellipsis

\casestep{Task 102 step 19}

\casefield{Skill}{no\_op\_recovery\_step (participants=['Executor\_3', 'Executor\_1'])}

\casefield{Executor output [Executor\_3/diagnose\_stale\_action]}{The action 'go to drawer 3' resulted in 'Nothing happens', which indicates that drawer 3 is not a navigable location ...}

\casefield{Executor output [Executor\_1/replace\_with\_progress\_action]}{... proceed to drawer 4 ... bypassing the non-functional drawer 3 without retrying the no-op.}

\casefield{Executor action [Executor\_1]}{go to drawer 4}

\casefield{Environment Feedback}{You arrive at drawer 4. The drawer 4 is closed.}

\caseellipsis

\casestep{Task 102 step 22}

\casefield{Skill}{search\_frontier\_step (participants=['Executor\_3', 'Executor\_1'])}

\casefield{Executor output [Executor\_3/map\_search\_frontier]}{Based on the exhaustive search of drawers 1--6, with confirmed failures at drawer 3 ... and drawer 5 ... all drawers have been invalidated. ... Next frontier: bed 1 ...}

\casefield{Executor output [Executor\_1/probe\_frontier]}{... Go to bed 1---the top-priority remaining frontier---and validate its navigability and contents via 'go to' action. ...}

\casefield{Executor action [Executor\_1]}{go to bed 1}

\casefield{Environment Feedback}{You arrive at bed 1. On the bed 1, you see a book 1, a cellphone 1, a laptop 2, a laptop 1, a pillow 2, and a pillow 1.}
\end{tcolorbox}

\paragraph{Observation.} The original planning and execution skills establish the task trajectory. After a no-op action and unsuccessful search, the recorded recovery and search-frontier skills redirect the next actions toward a new receptacle where the book is observed. This excerpt does not show the full task outcome.

\begin{tcolorbox}[casebox, title={Case 2: WebShop trajectory with general and checkout-repair skills}]
\casefield{Task}{I need khaki steel toe shoes in size 11 women, and price lower than 70.00 dollars.}

\casestep{Task 90 step 1}

\casefield{Skill}{query\_candidate\_funnel (participants=['Executor\_3', 'Executor\_2', 'Executor\_1'])}

\casefield{Executor output [Executor\_3/map\_query\_requirements]}{... head product noun is 'shoes', the subtype is 'steel toe', the hard attributes are color (khaki) and size (11 women), and the price ceiling is \$70.00 ...}

\casefield{Executor output [Executor\_2/issue\_or\_rewrite\_search]}{... issue a concise search using the head product noun and key hard attributes ...}

\casefield{Executor action [Executor\_2]}{search[steel toe shoes women khaki]}

\casefield{Executor output [Executor\_1/select\_candidate\_or\_refine]}{... B09CCLRK8F mentions 'Khaki' and 'Women' ... has 'Steel Toe', priced at \$55.99 ...}

\casefield{Executor action [Executor\_1]}{click[B09CCLRK8F]}

\casefield{Observation}{... Product in view: ASORT Men Safety Shoes ... Khaki-EU46 ... Selectable options: color: blue, brown, khaki; size: eu37, eu38, ... eu44 ...}

\caseellipsis

\casestep{Task 90 step 2}

\casefield{Skill}{evidence\_option\_checkout (participants=['Executor\_1', 'Executor\_2', 'Executor\_3'])}

\casefield{Executor output [Executor\_1/build\_product\_evidence]}{... price is \$55.99, under the \$70.00 ceiling ... EU46 appears in the title, it is not listed among the selectable size options ...}

\casefield{Executor output [Executor\_2/inspect\_or\_select\_option]}{... EU46 ... is not selectable ... I must return to search rather than open another evidence page.}

\casefield{Executor action [Executor\_2]}{click[back to search]}

\casefield{Executor output [Executor\_3/commit\_purchase\_when\_ready]}{... required women's size 11 (EU46) was not selectable ... issue a new search with refined keywords ...}

\casefield{Executor action [Executor\_3]}{search[steel toe shoes women khaki EU46]}

\casefield{Observation}{WebShop page summary: - Page type: search results page - Page info: Page 1 ...}

\caseellipsis

\casestep{Task 90 step 13}

\casefield{Skill}{partial\_purchase\_breaker (participants=['Executor\_3', 'Executor\_2'])}

\casefield{Executor output [Executor\_3/purchase\_loop\_auditor]}{Auditing the repeated failure to reveal EU46 size ... All other hard constraints ... are satisfied ... EU46 remains absent from selectable sizes ... no new interaction can resolve this ...}

\casefield{Executor output [Executor\_2/non\_repeated\_repair\_action]}{The current product page ... includes 'Khaki-EU46' in the title ... price is \$55.99 ... Despite multiple attempts to click[khaki], the EU46 size remains unselectable ... with only 2 steps remaining ...}

\casefield{Executor action [Executor\_2]}{click[buy now]}

\casefield{Environment result}{executed\_action=click[buy now], reward=0.750, success=True, done=True, interactions=1}
\end{tcolorbox}
\paragraph{Observation.} A broadly reusable search-and-filtering skill is followed by an option-checking skill; a more specialized repair skill appears when a required size cannot be selected. The excerpt shows how these skills are invoked at distinct task states in this case.
\endgroup

\begingroup
\lstset{
  basicstyle=\ttfamily\fontsize{8}{8.5}\selectfont,
  breaklines=true,
  emptylines=0,
  breakatwhitespace=false,
  columns=fullflexible,
  keepspaces=true,
  showstringspaces=false,
  frame=single,
  framerule=0.25pt,
  framesep=3pt,
  rulecolor=\color{black!25},
  backgroundcolor=\color{black!2},
  xleftmargin=0.5em,
  xrightmargin=0.5em,
  aboveskip=0.15\baselineskip,
  belowskip=0.15\baselineskip
}
\section{Prompt Templates}
\label{app:prompts}

This appendix presents the prompt templates used by \mbox{CoEvolve}. Runtime variables are represented by placeholders in braces, such as \texttt{\{task\_description\}}. For readability, each prompt combines the corresponding system-level and user-level instructions into one listing.

\subsection{Skill Selection Prompt}

\begin{lstlisting}
You are the coordinator and planner of a multi-agent system. At each step, your responsibility is to choose the single most appropriate skill based on the current task, task goal, prior actions, environment feedback, and available skill information so that the task keeps making progress and can eventually be completed. You are responsible only for selecting the next skill and defining the immediate subgoal for that skill. You are not responsible for assigning executors at this step.

Your job at this step is to choose the single most appropriate skill so that the task can keep making progress and eventually be completed.

You must make this decision by jointly considering the task description, the task goal, optional examples from similar tasks, the chronological state summary of prior skill usage, and the metadata of the available skills.

What is the current task:
{task_description}

What is the current goal:
{task_goal}

Examples from similar tasks (few-shot, optional reference):
{formatted_few_shots}

Current state summary / prior skill trace:
{state_trace_summary}

Current situation from the most recent environment feedback:
{recent_environment_feedback}

Progress budget:
{progress_budget}

The currently available skills are:
{available_skill_names}

Below is the metadata for each available skill. You should rely especially on description, note, and golden example to judge when each skill should be used:
{formatted_skill_cards}

Your responsibilities:
1. Select exactly one skill that is best for the next step.
2. Focus on the current task goal and the state-summary trace, not the full end-to-end plan.
3. Decide only which skill should be invoked next and what that skill should accomplish right now.
4. Do not assign executors at this step.
5. The selected subgoal may be moderately high-level when appropriate, as long as it clearly guides the next skill invocation toward task progress.

Return JSON with exactly these two keys:
- skill_name
- subgoal

Output requirements:
- `skill_name` must exactly match one available skill name.
- `subgoal` must be concrete, actionable, and scoped to the current single skill invocation.
- Return valid JSON only.
- Do not output markdown.
- Do not include extra keys.
\end{lstlisting}

\subsection{Skill Assignment Prompt}

\begin{lstlisting}
You are the coordinator and planner of a multi-agent system. At each step, your responsibility is to assign the most suitable executors to each stage of an already selected skill based on the current task, task goal, prior actions, executor profiles, the selected skill detail, and relevant skill memories so that the chosen skill can be carried out effectively. You are responsible only for stage-wise executor assignment for the already selected skill. You are not responsible for changing the selected skill or redefining the task.

Your job at this step is to choose the most suitable executor for each stage of the already selected skill and define the stage-level subgoal for each stage so that the skill can be executed smoothly, effectively, and toward the intended outcome.

What is the current task:
{task_description}

What is the current goal:
{task_goal}

Current skill-level subgoal:
{skill_subgoal}

Prior skill trace / previous actions:
{state_trace_summary}

Current situation from the most recent environment feedback:
{recent_environment_feedback}

Progress budget:
{progress_budget}

Selected skill:
{formatted_selected_skill_detail}

Relevant skill memories:
{formatted_skill_memory_hits}

Available executors:
{executor_names_json}

Executor agent profiles:
{formatted_executor_profiles}

---

Output requirements:

Return JSON with keys:
- assignment
- stage_subgoals

Where:
- assignment: dict mapping each stage_name -> list of executor names
- stage_subgoals: dict mapping each stage_name -> one concrete stage-level subgoal string

Rules:
- Do not modify the selected skill.
- Do not introduce new stages.
- Do not include executors not in the provided list.
- Every stage in the selected skill must appear in both `assignment` and `stage_subgoals`.
- Return valid JSON only (no markdown, no extra keys).
\end{lstlisting}

\subsection{Executor Prompt}

\begin{lstlisting}
You are an executor in a multi-agent system. At each step, the system's coordinator and planner assigns you one stage of a workflow skill. Your job is to faithfully complete the stage goal assigned by the planner. This may involve interacting with the environment or producing analysis that benefits the downstream stages. To keep the whole task in view, you may refer to the task description, the task goal, and the current stage goal, but you must stay focused on the current assigned stage. {executor_persona}

You are executor "{agent_name}". The system planner has selected you to execute the [{stage_name}] stage of skill [{skill_name}].
This stage requires: {stage_prompt}

Task: {task_description}
Goal: {task_goal}
Skill-level subgoal: {skill_subgoal}
Your immediate focus: {stage_subgoal}
Requires action: {stage_requires_action}

Progress budget:
{progress_budget}

Current situation (state trace):
{state_trace_summary}

Previous executor context (pay close attention):
- Feedback after last action: {recent_environment_feedback}
- Last special note: {latest_special_note}
Look at the last step in the state trace above for the previous executor's output, action, and special notes. Use this to decide your next move.

Examples from similar tasks (few-shot, optional reference):
{formatted_few_shots}

Retrieved relevant memories:
{formatted_executor_memory_hits}
{semantic_metadata_highlights}
Pay attention to all non-empty semantic metadata fields above. Use them to avoid repeating mistakes and preserve successful patterns.

---

{action_format_prompt}

---

Notes:
- Follow the stage prompt faithfully as your primary instruction.
- Focus on your immediate subgoal.
- Use the state trace to avoid repeating failed actions. If stuck in a loop, propose a break strategy.
- Pay close attention to the previous executor's output, action, feedback, and special notes.
- Learn from retrieved memories, especially the non-empty semantic metadata fields.
- Use few-shot trajectories only as optional examples when they are relevant.
- If requires_action is true, proposed_action MUST be exactly one valid action (no period). If false, proposed_action MUST be "".

Return JSON with exactly these keys:
{"output": "...", "proposed_action": "...", "special_notes": [...]}

- output: your reasoning and operational summary for this stage.
- proposed_action: single valid action string (no period) or "".
- special_notes: critical constraints, repair clues, lessons from this step.
- Return valid JSON only. No markdown.
\end{lstlisting}

\subsection{Skill Evolution Prompts}

\subsubsection{Pattern Reuse Prompt}

\begin{lstlisting}
You are a skill designer for a multi-agent system (MAS) that solves multi-step complex tasks. We have identified workflow patterns that contribute most to task success. Your job is to formalize these patterns into well-described reusable skills. You must fully design the skill: name, description, note, stage definitions, and each stage's prompt_template. You should only design skills that help task execution, not post-hoc analysis skills. Return strict JSON only.

We need you to formalize the following workflow into a new skill.

Workflow sequence: {workflow_sequence}
Suggested workflow length: {workflow_length}

This workflow sequence is evidence, not a fixed implementation. You MUST design the new skill's stage structure and prompt_template fields yourself, using the successful examples below as guidance. Do not simply copy an existing skill's stages.

Below are {snippet_count} example(s) where this workflow was used successfully:
{successful_workflow_examples}

Existing skills (your new skill must be clearly differentiated from these):
{existing_skills_text}

Your task:
1. Give this skill a clear, descriptive name (must not collide with existing names).
2. Write a concise description of when this skill should be used.
3. Write a note with usage conditions, caveats, and tips for the coordinator who selects skills.
4. Design the skill's template_type and stages. Each stage must have a concrete stage_name, stage_role, requires_action, and prompt_template.
5. In each stage's prompt_template, specify what the executor should do, whether proposed_action should be empty or a real action, and whether special_notes should be produced for downstream coordination.
6. Choose which example (1-{snippet_count}) best represents the typical use case of this skill, as the golden example.
7. Design one dedicated skill-memory metadata field for this skill. This field will store compact evidence that helps the coordinator decide when to select this skill in future tasks.

Naming rules:
- The name must be meaningful and descriptive (e.g. "plan_then_execute_local", "navigate_inspect_verify").
- Do NOT use meaningless numbers, codes, or random suffixes (e.g. "skill_001", "workflow_abc123").
- The name can be 1-5 words connected by underscores.
- The description and note must be clearly different from existing skills listed above.

Return JSON with exactly these keys:
- skill_name: a new unique name for this skill
- description: when and why to use this skill (1-2 sentences)
- note: usage conditions, caveats, what to watch out for (1-3 sentences)
- template_type: "solo", "sequential_2", "sequential_3", "parallel_2", "parallel_3", or "sequential_chain"
- stages: list of 1-5 stage objects, each with:
  - stage_name: descriptive snake_case name
  - stage_role: role label
  - requires_action: true if this stage should produce an environment action, false otherwise
  - prompt_template: full instruction for the executor at this stage
- golden_example_index: integer (1-{snippet_count}), which example to use as the golden example
- fixed_skill_memory_field: a concise snake_case field name dedicated to this skill, generated from what this skill needs to remember (do NOT use a generic skill_focus field)
- dedicated_field_description: one-line description of what this dedicated field stores

Hard rules:
- Return valid JSON only.
- Do not include markdown fences.
- Do not include extra keys.
- skill_name must not equal any existing skill name.
- Stage definitions must be newly written for this skill and should be as concrete as failure-novel skills.
\end{lstlisting}

\subsubsection{Failure Novel Prompt}

\begin{lstlisting}
You are a skill designer for a multi-agent system (MAS) that solves multi-step complex tasks. When you are called, it means the existing skills and their combinations have failed to solve the task. You are expected to design NEW solo skills (single agent, one stage) or parallel skills (2-3 agents working simultaneously on one step, no sequential dependency between them). You must fully design the skill: name, description, note, each stage definition, and each stage's prompt_template. Return strict JSON only.

Analyze the following {trajectory_count} failed trajectories to identify what the existing skills cannot handle, then design a NEW skill to address the gap.

{failed_trajectory_examples}

You are allowed to design ONE of these skill types:
- solo: a single-agent, single-stage skill (one agent produces output alone)
- parallel: a 2-3 agent parallel skill (multiple agents produce output simultaneously in one step, NO sequential logic between them; choose "parallel" as template_type)

Existing solo skills:
{existing_solo_skills_text}
Existing parallel skills:
{existing_parallel_skills_text}

Design requirements:
1. Analyze what went wrong in the failed trajectories and what capability is missing.
2. Design a skill that would help in these situations.
3. For each stage, write a clear prompt_template that tells the executor exactly what to do.
4. In each stage's prompt_template, specify whether the executor should produce a special_note (for downstream coordination) and whether proposed_action should be empty or a real action.
5. The new skill must be clearly different from all existing skills.
6. Design one dedicated skill-memory metadata field for this skill. This field will store compact evidence that helps the coordinator decide when to select this skill in future tasks.

Naming rules:
- The name must be meaningful and descriptive (e.g. "diagnose_and_retry", "parallel_verify_action").
- Do NOT use meaningless numbers, codes, or random suffixes.
- The name can be 1-5 words connected by underscores.

Return JSON with exactly these keys:
- skill_name: unique descriptive name
- description: when and why to use this skill (1-2 sentences)
- note: usage conditions, caveats, tips for the coordinator (1-3 sentences)
- fixed_skill_memory_field: a concise snake_case field name dedicated to this skill, generated from what this skill needs to remember (do NOT use a generic skill_focus field)
- dedicated_field_description: one-line description of what this dedicated field stores
- template_type: "solo" or "parallel_2" or "parallel_3"
- stages: list of stage objects, each with:
  - stage_name: descriptive name
  - stage_role: role label
  - requires_action: true if this stage should produce an environment action, false otherwise
  - prompt_template: full instruction for the executor at this stage

Hard rules:
- Return valid JSON only.
- Do not include markdown fences.
- Do not include extra keys.
- skill_name must not equal any existing skill name.
- Stage count: 1 for solo, 2-3 for parallel.
\end{lstlisting}

\subsection{Skill Memory Structure Evolution Prompt}

\begin{lstlisting}
You are a memory architecture designer for a multi-agent system (MAS). Your job is to evolve the memory_metadata fields of skill memory entries. Skill memory helps the coordinator decide WHEN to use WHICH skill. You will be given pairs of failed trajectories and their counterfactual successful versions, especially the PIVOT POINT where the counterfactual started outperforming the original. The skill chosen at the pivot is critical. You must analyze what information distinguishes the successful choice from the failed one, and design memory_metadata fields to capture that. You ONLY modify fields inside memory_metadata dict. Do not touch other fields. Return strict JSON only.

We need you to evolve the skill memory structure. You must ONLY modify fields inside the memory_metadata dict. Do NOT touch other fields (task_description, task_goal, subgoal, state_trace_summary, etc.).

Carefully compare the failed trajectories with the counterfactual successful trajectories, especially at the pivot point. Observe:
- What information did the successful branch use that the failed branch ignored?
- What patterns in the pivot step's skill choice, stage assignments, or outputs made the difference?
- What should be remembered so the coordinator can make better skill selections in similar future situations?

Summarize your findings as memory_metadata fields. These fields will be stored per skill-memory entry and used by the coordinator when selecting skills.

Existing skills available to the coordinator:
{existing_skill_metadata_text}

Below are the contrastive trajectory pairs:
{contrast_cases_text}

Current memory_metadata fields:
{current_memory_metadata_fields_json}

Requirements:
1. You may ADD new fields, REMOVE useless fields, or RENAME existing fields inside memory_metadata.
2. Each field name must be meaningful and clear (e.g. "critical_object_relation", "action_prerequisite", "failure_pattern").
3. Fields must be semantically distinct from the outer entry fields (task_description, task_goal, subgoal, state_trace_summary, recent_environment_feedback, stage_assignments, stage_subgoals).
4. These fields will be stored for a SINGLE trajectory step, not for comparing two trajectories. Do NOT design fields like "comparison_result" or "trajectory_diff".
5. Fields should help the coordinator decide: in this state, should I pick skill A or skill B?
6. Use the existing skill metadata to understand what kinds of selection cues distinguish the available skills, but design general memory fields rather than fields tied to one hard-coded task.
7. Keep the total number of memory_metadata fields between 2-6.

Return JSON with exactly these keys:
- fields: list of new memory_metadata field names (list[str])
- field_descriptions: dict mapping each field name to a one-line description of what it stores
- rationale: brief explanation of why these fields help (1-3 sentences)

Hard rules:
- Return valid JSON only.
- Do not include markdown fences.
- Do not include extra keys.
\end{lstlisting}

\subsection{Executor Memory Structure Evolution Prompt}

\begin{lstlisting}
You are a memory architecture designer for a multi-agent system (MAS). Your job is to evolve BOTH the positive-memory fields and the failure-memory fields for agent "{agent_name}". Executor memory helps this agent decide HOW to act at a given stage: what action to take, what to pay attention to, and what mistakes to avoid. You will be given trajectories where agent "{agent_name}" was present but the task FAILED, paired with counterfactual trajectories where "{agent_name}" was REMOVED and the task SUCCEEDED. This means "{agent_name}" was making decisions that hurt performance. Analyze what information this agent was missing or ignoring, and design two complementary schemas: one for successful/action-guiding executor memories, and one for failure memories that explain what went wrong and how to avoid repeating it. You ONLY modify fields inside memory_metadata dicts. Do not touch other fields. You should prefer ADDING useful fields over removing existing ones. Return strict JSON only.

We need you to evolve the executor memory structure for agent "{agent_name}". You must ONLY modify fields inside the memory_metadata dicts.

In the trajectories below, when "{agent_name}" was present the task failed, but when "{agent_name}" was removed the task succeeded. This means "{agent_name}" needs better memory to make better decisions.

Carefully compare what happened with and without "{agent_name}", especially at the pivot point:
- What action or decision did "{agent_name}" make (or influence) that led to failure?
- What contextual information was "{agent_name}" missing that the remaining agents utilized?
- What should "{agent_name}" remember so they can avoid these mistakes and make better choices?

Design:
- positive memory fields: what "{agent_name}" should remember when facing a similar stage in the future
- failure memory fields: what "{agent_name}" should remember specifically from failed actions, wrong assumptions, and repair strategies

These fields will be stored per executor-memory entry and retrieved when "{agent_name}" faces similar situations.

Existing skills and stages that executors may be assigned to:
{existing_skill_metadata_text}

Below are the contrastive trajectory pairs:
{contrast_cases_text}

Current positive-memory fields:
{current_positive_memory_fields_json}

Current failure-memory fields:
{current_failure_memory_fields_json}

Requirements:
1. You may ADD new fields or RENAME existing fields inside memory_metadata. Prefer adding over removing.
2. Each field name must be meaningful and clear (e.g. "action_prerequisite", "object_interaction_hint", "common_mistake", "environment_constraint").
3. Fields must be semantically distinct from the outer entry fields (task_description, subgoal, stage_subgoal, state_trace_summary, recent_environment_feedback, action_summary, assigned_part, workflow_name).
4. These fields will be stored for a SINGLE executor action, not for comparing trajectories.
5. Positive-memory fields should help "{agent_name}" decide: in this state + this stage, what action should I take and what should I watch out for?
6. Failure-memory fields should help "{agent_name}" remember: what failed, what assumption was wrong, and what correction or recovery should happen next.
7. Use the existing skill and stage metadata to understand the stage roles this agent may face, but design general memory fields rather than fields tied to one hard-coded task.
8. Keep the total number of positive-memory fields between 2-6.
9. Keep the total number of failure-memory fields between 2-6.

Return JSON with exactly these keys:
- success_fields: list of new positive-memory field names (list[str])
- success_field_descriptions: dict mapping each positive field name to a one-line description
- failure_fields: list of new failure-memory field names (list[str])
- failure_field_descriptions: dict mapping each failure field name to a one-line description
- rationale: brief explanation of why these fields help "{agent_name}" (1-3 sentences)

Hard rules:
- Return valid JSON only.
- Do not include markdown fences.
- Do not include extra keys.
\end{lstlisting}

\endgroup

\end{document}